%% file: main.tex
\documentclass[letterpaper, 10 pt, conference]{ieeeconf}  

\IEEEoverridecommandlockouts                              

\usepackage{stfloats} 

\usepackage{graphics}
\usepackage{caption}
\usepackage{epsfig} 
\usepackage{mathptmx} 
\usepackage{times} 
\usepackage{amsmath} 
\usepackage{amssymb}  

\usepackage{hyperref}
\usepackage{algorithm}
\usepackage{algpseudocode}
\usepackage{multirow}
\usepackage[table]{xcolor}
\usepackage{lipsum}
\usepackage{comment} 
\usepackage{pifont}
\usepackage{listings}
\usepackage{bm}
\usepackage{footmisc}
\usepackage{balance}
\usepackage{multirow}

\usepackage[backend=biber, style=ieee, maxbibnames=19, minbibnames=19]{biblatex}
\title{\LARGE \bf
Puppeteer Your Robot: \\ Augmented Reality Leader-Follower Teleoperation

}

\author{Jonne van Haastregt*, Michael C. Welle*, Yuchong Zhang, Danica Kragic
\thanks{*These authors contributed equally (listed in alphabetical order).}
\thanks{KTH Royal Institute of Technology Stockholm, Sweden, {\it\small \{ 
        jmvh, mwelle, yuchongz, dani\}@kth.se}. }
}

\begin{document}

\maketitle

\thispagestyle{empty}
\pagestyle{empty}

\begin{abstract}
High-quality demonstrations are necessary when learning complex and challenging manipulation tasks. In this work, we introduce an approach to \textit{puppeteer} a robot by controlling a virtual robot in an augmented reality setting. Our system allows for retaining the advantages of being intuitive from a physical leader-follower side while avoiding the unnecessary use of expensive physical setup. In addition, the user is endowed with additional information using augmented reality. We validate our system with a pilot study $n=10$ on a block stacking and rice scooping tasks where the majority rates the system favorably. Oculus App and corresponding ROS code are available on the project website\footnote{\url{https://ar-puppeteer.github.io/}}.

\end{abstract}

\input{includes/intro}
\input{includes/rw}
\input{includes/framework}

\input{includes/user}
\input{includes/discussion}


\printbibliography

\end{document}

%% file: includes/intro.tex
\section{Introduction}

Recent advances in the field of robot learning from demonstration and behavior cloning have enabled complex manipulation tasks such as $6$Dof mug flipping,  sauce pouring and spreading~\cite{chi2023diffusion}, cooking shrimp and wiping wine~\cite{fu2024mobile}, or serving rice and opening bottles using a bottle opener~\cite{ingelhag2024robotic}.
A fundamental building block that enables these complex impressive manipulation tasks are high-quality expert demonstrations used for learning. As the learning methods are conditioned on the camera input to diffuse the next action~\cite{chi2023diffusion}, kinesthetic teaching approaches where the human is  guiding the robot are not applicable. This raises the need of intuitive and reactive teleoperation systems that facilitate high-quality demonstrations obtained from a wide range of operators.
To this end, the authors in ~\cite{chi2023diffusion} used a SpaceMouse to control a single Franka Panda arm, while~\cite{ingelhag2024robotic} related the velocities of an Occulus Quest $2$ controller to the cartesian velocities of the robot end-effector deploying the MetaLabs app Quest2ROS~\cite{welle2024quest2ros}. The authors in~\cite{fu2024mobile} designed a leader-follower teleoperation system that allows the operator to physically move the leader arms while in full view of the follower arms enabling intuitive bimanual manipulation. Such a \textit{puppeteer} setup enables complex teleoperation in $6$D, as leading the arms directly is very intuitive. However, in order to build such a setup the hardware has to be duplicated as the leader and follower arms have to be kinematicly very similar to make the teleoperation seamless, which doubles the hardware needed for such a system.

\begin{figure}[t]
    \centering
    \includegraphics[width=\linewidth]{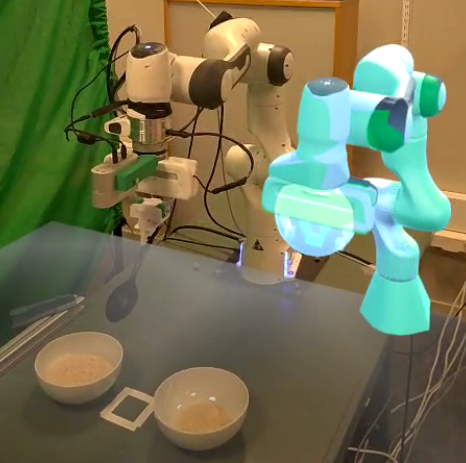}
    \caption{Our method uses an augmented reality (AR) approach to enable puppeteering without duplication of the real hardware. A virtual robot (right) is used as  leader and the real robot (left) acts as follower.}
    \label{fig:pup_front}
\end{figure}

\begin{figure*}[t]
    \centering
    \includegraphics[width=\textwidth]{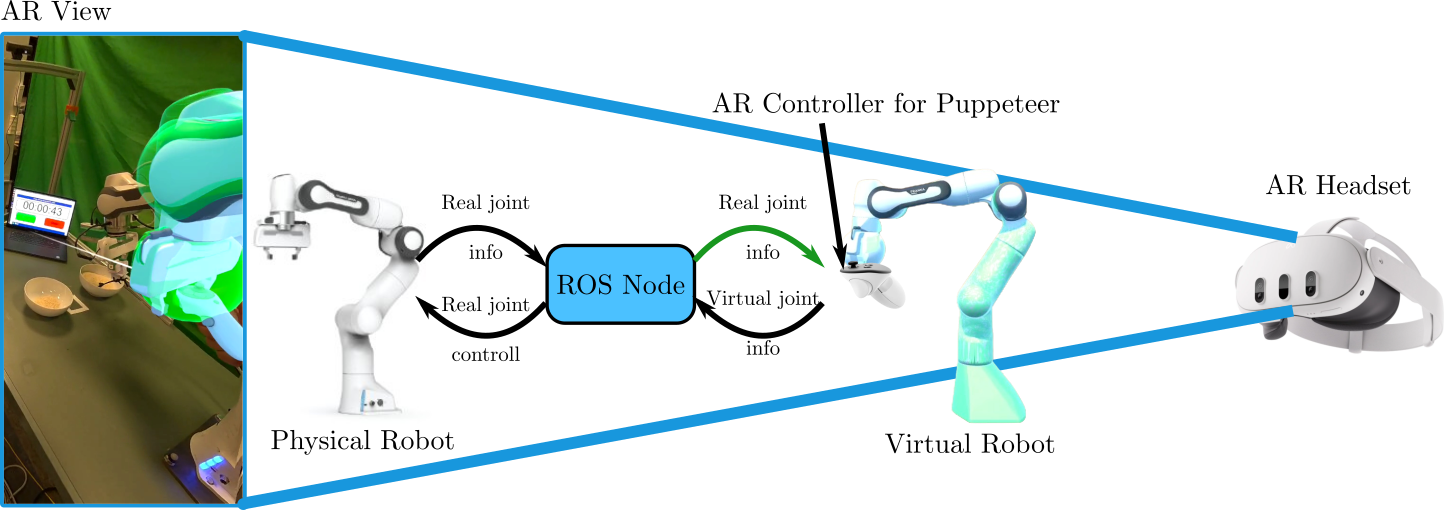}
    \caption{An overview of our AR puppeteer system. The Oculus projects the virtual robot into the line of sight of the user, who then uses the controller to \textit{grasp} the virtual robot at the end-effector and subsequently move it around. The resulting virtual joint positions are sent to a ROS Node and translated into suitable control signals for the real robot. Furthermore, the real robot's current joint position is relayed back to the Occulus over the same ROS node and enables the visualization between the wanted and current position of the robotic arm by visualizing the current position as a transparent green version of the robot. The user sees the composited augmented reality view (left) composed of the virtual robot, and the real robot.}
    \label{fig:overview}
\end{figure*}

In this work, we introduce a leader-follower teleoperation system that uses Augmented Reality (AR) to realize the leader arm, removing the need to duplicate the hardware but still provide an intuitive \textit{puppeteer} setup for the user. 
AR is recognized as a pervasive technique that superimposes virtual elements onto reality \cite{zhang2023see}, thereby enhancing information richness \cite{zhang2021supporting} and facilitating 3D visualization analysis \cite{zhang2023playing}. 
Our system enables the user to spawn a virtual Franka panda robot at a user-specified location using the Oculus Quest $3$ AR (passthrough) mode as shown in Fig. \ref{fig:pup_front}. 
This virtual manipulator serves as the leader robot in our setup, and to move the robot the user has to \textit{grasp} it on its end-effector using the Oculus Quest $3$ controller. 
Once the robot is gripped, the user can move the virtual end-effector around while the virtual robot is adhering to the kinematic constraints a real robot would impose - similar to a physical leader robotic arm. The resulting joint positions are then streamed to a ROS-TCP endpoint where a low-level joint controller mimics the joints of the virtual robot in reality. 
Furthermore, the actual position of the real robot, the follower, is relayed back to the AR headset and visualized as a (green) semi-transparent rendering indicating the \textit{delay} of the follower robot to the controlled robot instantly to the user. 
An overview of this setup, including a resulting first-person AR view is shown in Fig. \ref{fig:overview}.

We evaluated this setup by running a pilot study with $n=10$ participants that completed two manipulation tasks of different complexity, namely block stacking and rice scooping showing that this system enables all users to complete such tasks to various degrees of proficiency.

In detail, our contributions are:
\begin{itemize}
    \item A novel augmented Rreality based leader-follower teleoperation system for the Franka Panda available on Metas AppLabs (Puppeteer Franka)\footnote{\url{https://ar-puppeteer.github.io/\#app}}, and 
    \item An empirical evaluation demonstrating that our proposed system excels in user experience, usability, and preference.
\end{itemize}

%% file: includes/rw.tex
\section{Background \& Related Work}

\subsection{Puppeteering in Robotics}
Puppeteering has been used in robotics to control robotic systems in an intuitive and human-like manner. Early work focused on manual control interfaces, where operators directly manipulated robotic limbs using mechanical linkages or teleoperation systems. In 2018, Holland et al. \cite{holland2018visual} compared using kinematics and inverse kinematics to using the puppeteer’s movement angles in puppeteering robots, showing that the kinematic approach was better for robots with more degrees of freedom. Slyper et al. \cite{slyper2015mirror} developed a "Mirror Puppeteering" system that allows users to easily create gestures for robotic toys, custom robots, and virtual characters by moving the robot’s limbs in front of a webcam. Aravind et al. \cite{aravind2015automated} presented a fully automated marionette theatre featuring robot manipulators with sixteen degrees of freedom, connected via a WIFI network to a master controller, including a user-friendly interface for recording puppet motions in sync with audio and video. However, there has been limited research focused on controlling one real robot to enable another real robot to imitate its actions. A similar study was conducted by L\'{u}\v{c}ny et al \cite{luvcny2023robot},  where a custom model for a humanoid robot was constructed using pre-trained self-supervised models. The robot learns to detect its own body’s 3D pose by leveraging features extracted from visual inputs and predefined posture models through self-exploration in front of a mirror.

\subsection{Augmented Reality in Robotics}
AR is used in robotics due to its enhanced visualization and interaction capabilities \cite{makhataeva2020augmented}. In contrast to purely VR-based approaches~\cite{moletta2023virtual}, AR technologies overlay digital information onto the physical world, providing operators with real-time feedback and intuitive control interfaces. In robotics, there are several examples of research deploying AR techniques in teleoperation. Milgram et al. \cite{milgram1993applications} explored the director/agent (D/A) metaphor in telerobotic interaction, emphasizing AR as a tool to facilitate human-robot synergy. It introduced the ARGOS (Augmented Reality through Graphic Overlays on Stereovideo) system, which used overlaid virtual elements. Quintero et al. \cite{quintero2018robot} harnessed AR to robot programming using a headset and a $7$-DOF robot arm which focused on interactive functions: trajectory specification, virtual motion previews, visualization of robot parameters, and online reprogramming during simulation and execution. The effectiveness of the AR interface was validated through a pilot study. Regarding vision-based navigation for autonomous humanoid robots, Mohareri et al. \cite{mohareri2011autonomous} developed a platform for human location positioning and navigation indoors and outdoors using mobile AR, where $3$D graphics and audio cues convey location information derived from smartphone camera input and pre-constructed image databases.

\subsection{Combining Puppeteering and AR in Robotics}
The integration of puppeteering techniques with AR represents a novel approach to robotic control, aiming to leverage the strengths of both methods, with an emphasis on cost reduction (utilizing virtual robots/agents) and enhancing safety. So far, research in this area has been rather limited. Beiczy et al. \cite{bejczy2020mixed} developed a mixed-reality teleoperation interface tailored for mobile manipulation tasks in sensitive production environments where human presence is restricted, enabling operators to control the robot’s end-effector trajectory based on virtual reality (VR) controller poses. Sakashita et al. \cite{sakashita2017you} used a VR telepresence system to remotely puppeteer a robot while offering remote puppet manipulation based on natural body and facial gestures. The system was demonstrated by user studies with both novice and experienced puppeteers. 

The work presented in this paper is, to the best of our knowledge, first-ever AR-based puppeteering system for teleoperating a real physical robot.

%% file: includes/framework.tex
\section{Puppeteer Your Robot}

An overview of our system is shown in Fig. \ref{fig:overview}:  the user views an integrated view from the virtual and real robot. Similarly to a physical leader-follower teleoperation setup,  the user can \textit{grasp} the end-effector of the leader, here virtual robot and move it around while the folllower robot is mirroring the leaders movements. To give the user an instantaneous sense of how well the follower is following the teleoperated commands, the position information of the real robot is relayed back to the AR headset and an additional rendering of a transparent robot is shown to indicate the delay between the leader and the follower.

\subsection{Setup and User Interface}

\begin{figure}
    \centering
    \includegraphics[width=\linewidth]{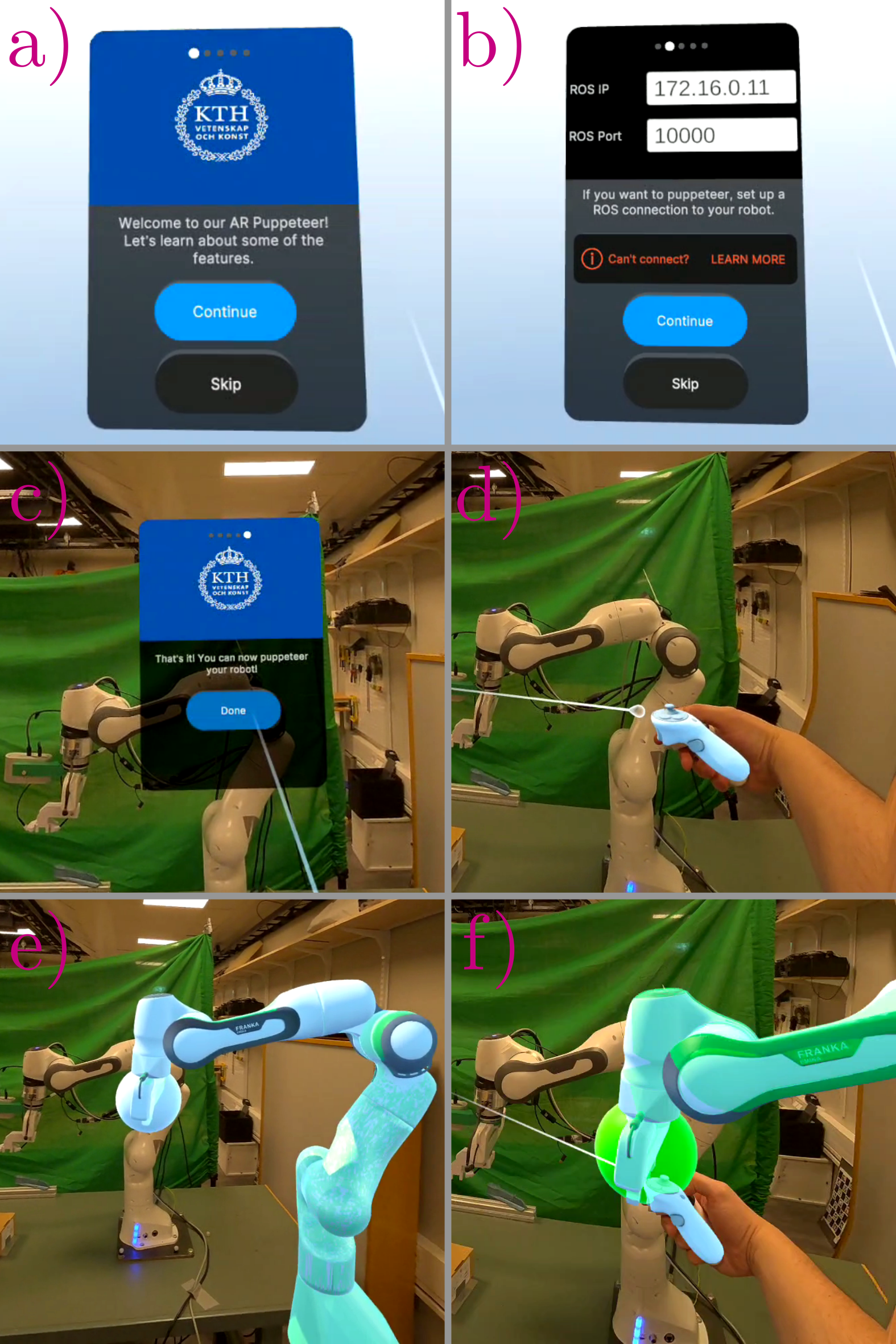}
    \caption{Setup of AR puppeteering in steps: a) Greeting screen, b) set ROS IP and Port to connect to, c) Setup completed screen, d) Position the controller to spawn the virtual robot, e) Virtual robot spawned, f) Pupeeterring engaged.}
    \label{fig:ui_setup}
\end{figure}

Fig.~\ref{fig:ui_setup} shows the app's user interface and how the AR puppeteering is set up. First, the app shows a greeting screen followed by the connection screen. In the connection screen, the user can set the ROS IP as well as the Port for the corresponding ROS-TCP connection\footnote{\fontsize{6}{8}\url{https://github.com/Unity-Technologies/ROS-TCP-Endpoint}}. The screen will persist until messages of the real robot state are perceived by the app from the corresponding ROS node.

After the connection is successfully established, the user is free to spawn the robot at a desired location. The user can then place the controller in an arbitrary position and orientation and press the "B" button on the right-hand controller to spawn in the robot, see Fig. \ref{fig:ui_setup} d \& e. The user is free to reposition the robot as many times as needed by pressing "B" again. Once the user is satisfied with the location of the virtual robot in the real scene, the puppeteering can be engaged by \textit{grasping} the virtual robot by the sphere around the end-effector. As long as the users controller is in this sphere and the "A" button on the right-hand controller is pressed the joint positions are relayed to the ROS node which translates them to actual joint commands using a PD-controller for the real robot. Furthermore, in this condition the gripper can be engaged by pressing the trigger button on the controller.

\subsection{AR Puppeteer Control}

\begin{figure}
    \centering
    \includegraphics[width=\linewidth]{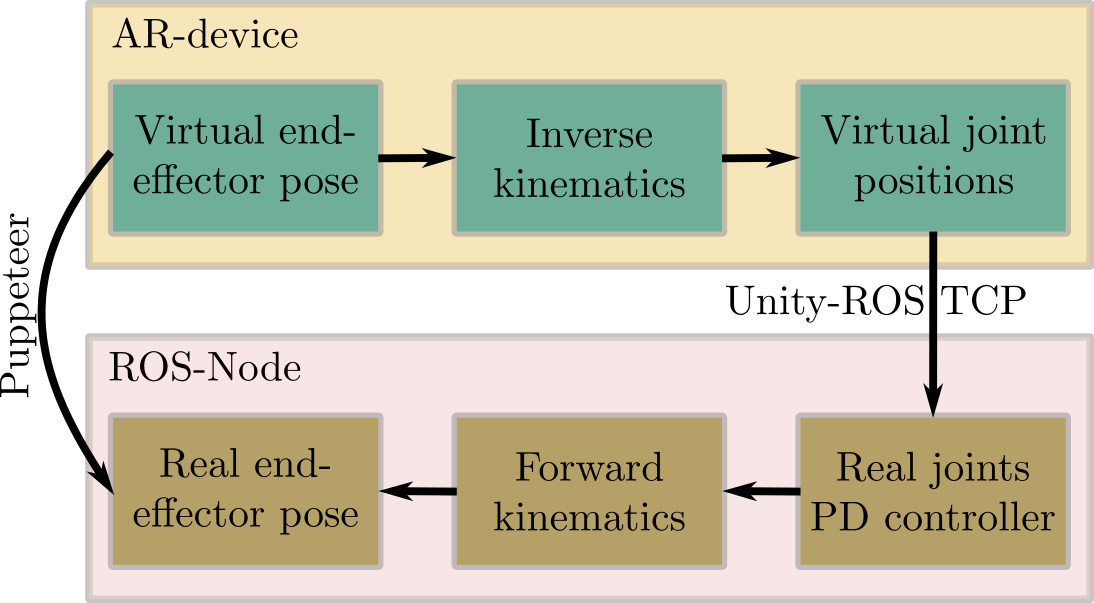}
    \caption{Overview of how the AR Puppeteer control is realized. The virtual joint positions are obtained via Inverse kinematics given the virtual end-effector pose. These joint positions are then published onto a ROS topic where a PD controller realizes them on the real robot, the kinematics of the robot then results in the real end-effector following the virtual one.}
    \label{fig:pup_controll}
\end{figure}

Fig. \ref{fig:pup_controll} shows a schematic overview of how the puppeteering is realized on a control level. 

To grasp the virtual robot, the user can hold the controller in the grasp area and press and hold 'A' button. The grasp area is a white sphere around the end-effector, which turns blue when the controller is within it and green when the grasping is active. 
A target for the end-effector is then set by following the motion of the controller. The target follows around the point where the user has grasped, as if there is a rigid link between the controller and the end-effector.

\begin{algorithm}
\caption{IK approximation}
\label{alg:ik}
\begin{algorithmic}
\State $\hat{q} \gets q$
\State $x_{error} \gets \infty$
\While {$x_{error} > \epsilon$}
\State $\hat{x} \gets T_{b-e}(\hat{q})$
\State $x_{error} \gets x_{target} - \hat{x}$ \label{algline:error}
\State $dq \gets J^+(\hat{q}) x_{error}$
\State $\hat{q} \gets \hat{q} + dq$
\EndWhile

\end{algorithmic}
\end{algorithm}

Using inverse kinematics, the desired joint targets of the virtual robot can be computed as shown in Algorithm~\ref{alg:ik}. We first the joint configuration $\hat{q}$ by setting it to the current joint configuration $q$. After initializing the task space error $x_{error}$, we iterate until the error is lower than a set tolerance $\epsilon$. We calculate $\hat{x}$ via the forward kinematics where $T_{b-e}(\hat{q})$ is the transformation from base to end-effector, next the error is updated in line \ref{algline:error}.
Given the new error, we calculate the change in joint configuration $dq$ where $J^+$ represent the pseudo-inverse of the Jacobian matrix.
Finally, the estimate of the joint configuration $\hat{q}$ is updated, and the next iteration starts until $x_{error}<=\epsilon$.

In some cases, for instance around singularities or when the difference between the current end-effector position and the target is too large, the method for computing the inverse kinematics may be unstable. When the method does not converge or exceeds velocity limits in its result, the new joint targets are not set. The grasp area will turn red to indicate that the grasping motions are not being applied anymore. Simply releasing the grasp and re-grasping will reset this behavior. The velocity limits are $2 m/s$ for the end-effector and $1 rad/s$ for each individual joint.

The virtual robot has to adhere to real-world physics and is simulated by applying torques resulting from a simple PD controller in every joint as in Eq.~\ref{eq:PD_AR}, where $k_p$ is the proportional gain and $k_d$ is the derivative gain.

\begin{equation}\label{eq:PD_AR}
    \tau = k_p(q - \hat{q}) - k_d\dot{q}
\end{equation}

Before providing the joint states of the virtual leader to the real-world follower, we check if the virtual robot state is close enough to the real robot state. If any joint has a difference bigger than $0.2$ rad, providing the joints will get locked until further action is taken to re-align the virtual robot to the real robot. The transparent robot rendering will turn red.

If the puppeteer can safely provide the virtual joint positions, they are sent over the Unity-ROS TCP connection to the real robot joint controller.
The controller receives the new target joint position $q$ with a frequency of approximately $50Hz$, while the main control loop of the Franka panda robot is running at $1000Hz$. To obtain a smooth trajectory, we deploy a low-pass filter to obtain the filtered position
\begin{equation}
\tilde{q_{t}} = (1-\alpha)\tilde{q_{t-1}} + \alpha q    
\end{equation}
where $\alpha$ is a smoothing factor between $0$ and $1$ - a low $\alpha$ results in a smoother trajectory but less accurate tracking of the desired $q$ while a high $\alpha$ results in more accurate tracking but jerky motions.

The needed joint torque is then obtained via a simple PD controller:
\begin{equation}
    \tau = K_p(\tilde{q_t}-q_t) + K_d\dot{q}
\end{equation}

where $K_p$ is the proportinal gain, $K_d$ is the derivative gain, and $\dot{q}$ is the current joint velocity.

\subsection{AR Puppeteer Usage}

The AR leader robot is endowed with a number of additional features that give the user more information about the tracking performance as well as safety features that avoid moving the virtual robot to fast.

\begin{figure}
    \centering
    \includegraphics[width=\linewidth]{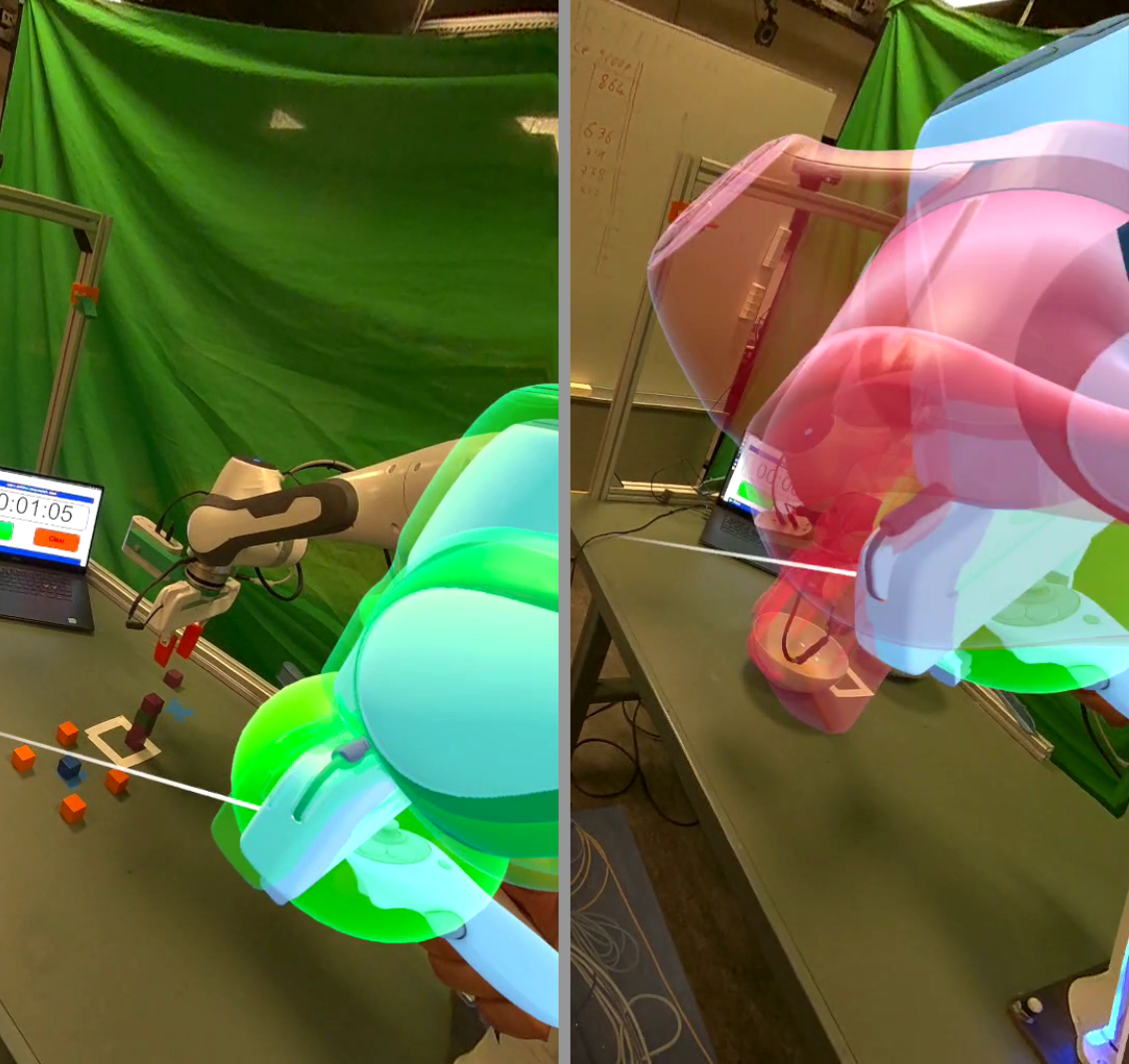}
    \caption{Visulisation of the state of the follower robot overlayed as a transparent robot to indicate the alignment between leader and follower in an intuitive way. Left: a green transparent robot indicates the \textit{delay} between the leader and follower. Right: a red transparent robot indicates that the follower diverged from the leader and that realignment is needed.}
    \label{fig:gr_shadow}
\end{figure}

One of the advantages of an AR-realized leader robot for a puppeteer setup is that information from the real robot can be sent back to the leader thus not being a simple one-way setup. The real robot (follower) joint positions are relayed back to the AR device and an additional transparent robot is displayed with the same base frame as the virtual robot indicating the current state of the real robot. The offset in the alignment between the transparent robot and the virtual robot gives the user instantaneous information on how the real robot is following the leader robot as shown in Fig.~\ref{fig:gr_shadow} (left). We color the transparent robot green when the divergence between leader and follower is under a certain threshold, indicating to the user also that the puppeteering is going as it should. If the follower diverges too far, if too high joint velocities on the virtual robot are detected, or a communication interruption is noted, the streaming is stopped for safety reasons and the transparent robot turns red, indicating that the virtual leader robot first has to be realigned with the follower (real robot), Fig.~\ref{fig:gr_shadow} (right). 
The user can perform the realignment by pressing the "X" button on the left-hand controller, which resets the virtual leader robot to the position of the real follower robot. After the realignment, the user is free to resume the puppeteering.

%% file: includes/user.tex
\section{Initial User Evaluation}

To empirically evaluate the efficiency of the puppeteering system and demonstrate its practical use among end users, we conducted a pilot study. The goal was also to assess the system’s usability and its potential for real-world application. We recruited $10$ participants (M=$7$, F=$3$, mean=$28.2$, SD=$2.86$) through the authors’ network at the local university. From a short demographic background investigation, we found that seven participants rated themselves as beginner in AR, while three possessed intermediate experience with AR before. Regarding prior exposure to robotics, the majority (eight out of ten) rated themselves as advanced. 
Before the study, all participants signed a consent form stating that no private data would be collected and that all results would be used solely for research purposes. Each participant completed a post-study questionnaire after their session. This questionnaire measured six key self-designed user experience (UX) and usability metrics (on a 7-point Likert scale: from strongly disagree to strongly agree) inspired by previous literature \cite{zhang2023playing,rubagotti2022perceived,hasan2007effects,adams2009multiple,bartneck2009measurement}: ease of use, intention to use, perceived safety, perceived ease of learning, perceived workload, and perceived likeability. Each study session for individual participants was approximately $30$ to $40$ minutes including a pre-training session. All data collection adhered to the ethical guidelines of the authors’ home universities.

\subsection{Procedure}
\begin{figure}
    \centering
    \includegraphics[width=\linewidth]{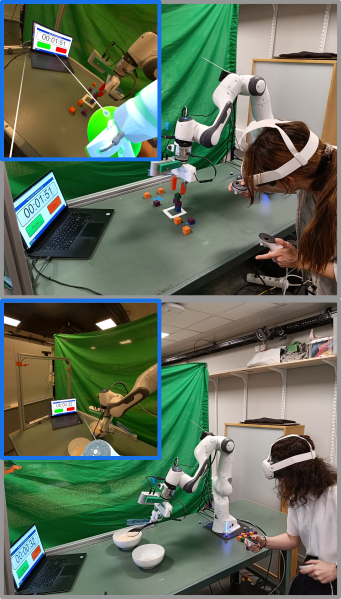}
    \caption{Snapschoots of the user study. On top, the block stacking task is shown, while the bottom depicts the rice scooping task. The blue border marks the corresponding AR view the participant's experience during the trials.}
    \label{fig:pup_users}
\end{figure}

To thoroughly evaluate the effectiveness of our puppeteering system in performing practical tasks, we designed and implemented two straightforward tasks in our study: cube stacking and rice scooping. For the cube stacking task, see  Fig.~\ref{fig:pup_users} (top),  participants were instructed to stack cubes one by one in a single stack placed in the white rectangle using the gripper on the end-effector of a physical robot arm via the puppeteer system. 
Participants were provided with a total of ten cubes and were tasked with stacking as many cubes as possible within a two-minute time limit. There were five cubes symmetrically placed on either side of the stacking area. The goal was to pick up these cubes and place them into the marked square area in the middle. The number of successfully stacked freestanding cubes at any point during a trial served as a key performance metric. 

In the rice scooping task, participants were required to scoop rice from one bowl to another using a spoon grasped by the end-effector, employing the same puppeteering manipulation technique as in the cube stacking. The task also had a two-minute time limit, and the weight of the rice (in grams) successfully scooped was recorded as another performance measure. 
As shown in Fig.~\ref{fig:pup_users} (bottom), the bowls used in the rice scooping task are identical, with one bowl filled with rice and the other used to receive the scooped rice being empty at the start of the trail. 

Prior to performing the tasks, each participant donned the headset and participated in a training session. This session allowed to practice controlling the physical robot via the virtual robot, helping the participants to become familiar with the system and reducing any novelty effects. All participants were given two trials for each task. 
After completing both tasks, participants were asked to fill out a brief questionnaire (available on the project website) to provide feedback on their user experience and the usability of our proposed system. This feedback is crucial for assessing the overall effectiveness and user satisfaction with the proposed puppeteering system. After finishing each trial, we either re-organized the locations of the cubes to return them to their initial positions or returned all the rice to its initial state to the original bowl to ensure the experimental conditions remained consistent.
For all participants, the smoothing factor $\alpha$ was set to $0.02$ and the control gains $K_p$ and $K_d$ were set to $[600,600,600,250,150,50,1], [50,50,20,20,20,10,1]$ for the joints $1-7$ respectively.

\subsection{Results}
After all participants completed the study, we collected the results and conducted post-study data analysis. There were in total eight metrics investigated: two performance metrics (the number of cubes stacked and the proportion of the rice scooped from the original bowl), and six UX metrics as aforementioned shown in Table~\ref{tab:results_all}.

\begin{figure}[t]
    \centering
    \includegraphics[width=\linewidth]{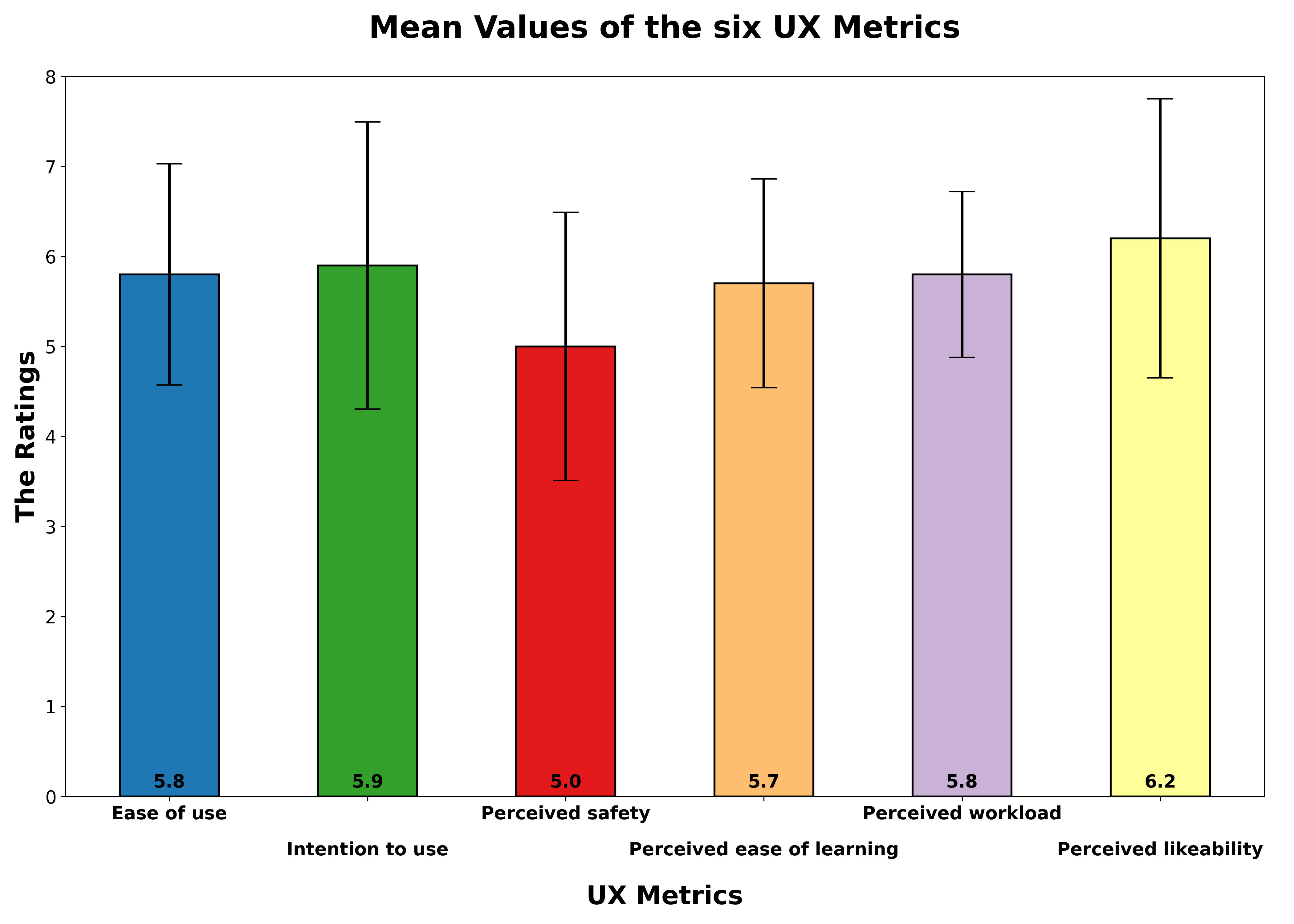}
    \caption{Mean values of the six UX metrics from our study.}
    \label{fig:ux_metrics}
\end{figure}

From the measurement of the performance metrics, we see that our AR-based puppeteering system was capable of supporting users in accomplishing practical tasks. In cube stacking, most participants successfully managed to stack over half of the cubes within the time limit, and most had an obvious performance amelioration in that second trial. The same phenomenon applied to the rice scooping task, where most participants effectively scooped a certain amount of rice via the puppeteering system and had a remarkable performance improvement in the second trial. Regarding the measured UX metrics, as shown in Fig.~\ref{fig:ux_metrics}, all metrics received satisfactory ratings, with the mean value of each scoring above 5 on a 7-point Likert scale. Generally, the reported perceived likeability had the highest ratings, while perceived safety had the lowest (the difference is mild). However, according to individual responses seen in Table~\ref{tab:results_all}, most participants reported a high level of perceived safety for our puppeteering system. Based on the collected results, we are confident in concluding that the proposed system is both empirically and practically advisable in terms of usability, UX, and user preference.

\begin{table*}
\centering
\resizebox{\textwidth}{!}{%
\begin{tabular}{|c|cccc|cccccc|}
\hline
\multirow{2}{*}{\#} & \multicolumn{4}{c|}{Performance Metrics}                                                                                                                                 & \multicolumn{6}{c|}{\begin{tabular}[c]{@{}c@{}}UX Metrics\\ Scale: 1-7 (Strongly Disagree -- Strongly Agree)\end{tabular}}                                                                                                                                                   \\ \cline{2-11} 
                    & \multicolumn{2}{c|}{\begin{tabular}[c]{@{}c@{}} No. of  cubes \\ \# stacked\end{tabular}} & \multicolumn{2}{c|}{\begin{tabular}[c]{@{}c@{}}Rice scooped\\ {[}g{]}\end{tabular}} & \multicolumn{1}{c|}{Ease of use} & \multicolumn{1}{c|}{Intention to use} & \multicolumn{1}{c|}{Perceived safety} & \multicolumn{1}{c|}{Perceived ease of learning} & \multicolumn{1}{c|}{Perceived workload} & Perceived likeability \\ \hline
1                   & \multicolumn{1}{c|}{7}                  & \multicolumn{1}{c|}{10}                  & \multicolumn{1}{c|}{158}                        & 264                       & \multicolumn{1}{c|}{6}           & \multicolumn{1}{c|}{6}                & \multicolumn{1}{c|}{4}                & \multicolumn{1}{c|}{5}                          & \multicolumn{1}{c|}{7}                  & 6                     \\ \hline
2                   & \multicolumn{1}{c|}{8}                  & \multicolumn{1}{c|}{7}                   & \multicolumn{1}{c|}{214}                        & 232                       & \multicolumn{1}{c|}{5}           & \multicolumn{1}{c|}{5}                & \multicolumn{1}{c|}{5}                & \multicolumn{1}{c|}{6}                          & \multicolumn{1}{c|}{5}                  & 6                     \\ \hline
3                   & \multicolumn{1}{c|}{4}                  & \multicolumn{1}{c|}{7}                   & \multicolumn{1}{c|}{74}                         & 36                        & \multicolumn{1}{c|}{6}           & \multicolumn{1}{c|}{5}                & \multicolumn{1}{c|}{5}                & \multicolumn{1}{c|}{6}                          & \multicolumn{1}{c|}{6}                  & 6                     \\ \hline
4                   & \multicolumn{1}{c|}{6}                  & \multicolumn{1}{c|}{7}                   & \multicolumn{1}{c|}{134}                        & 114                       & \multicolumn{1}{c|}{7}           & \multicolumn{1}{c|}{7}                & \multicolumn{1}{c|}{5}                & \multicolumn{1}{c|}{6}                          & \multicolumn{1}{c|}{5}                  & 7                     \\ \hline
5                   & \multicolumn{1}{c|}{4}                  & \multicolumn{1}{c|}{7}                   & \multicolumn{1}{c|}{90}                         & 178                       & \multicolumn{1}{c|}{6}           & \multicolumn{1}{c|}{6}                & \multicolumn{1}{c|}{7}                & \multicolumn{1}{c|}{5}                          & \multicolumn{1}{c|}{6}                  & 7                     \\ \hline
6                   & \multicolumn{1}{c|}{4}                  & \multicolumn{1}{c|}{5}                   & \multicolumn{1}{c|}{158}                        & 222                       & \multicolumn{1}{c|}{7}           & \multicolumn{1}{c|}{7}                & \multicolumn{1}{c|}{5}                & \multicolumn{1}{c|}{7}                          & \multicolumn{1}{c|}{7}                  & 7                     \\ \hline
7                   & \multicolumn{1}{c|}{7}                  & \multicolumn{1}{c|}{7}                   & \multicolumn{1}{c|}{140}                        & 152                       & \multicolumn{1}{c|}{6}           & \multicolumn{1}{c|}{7}                & \multicolumn{1}{c|}{6}                & \multicolumn{1}{c|}{7}                          & \multicolumn{1}{c|}{6}                  & 7                     \\ \hline
8                   & \multicolumn{1}{c|}{6}                  & \multicolumn{1}{c|}{5}                   & \multicolumn{1}{c|}{164}                        & 174                       & \multicolumn{1}{c|}{7}           & \multicolumn{1}{c|}{7}                & \multicolumn{1}{c|}{7}                & \multicolumn{1}{c|}{7}                          & \multicolumn{1}{c|}{6}                  & 7                     \\ \hline
9                   & \multicolumn{1}{c|}{2}                  & \multicolumn{1}{c|}{4}                   & \multicolumn{1}{c|}{48}                         & 54                        & \multicolumn{1}{c|}{3}           & \multicolumn{1}{c|}{2}                & \multicolumn{1}{c|}{2}                & \multicolumn{1}{c|}{4}                          & \multicolumn{1}{c|}{4}                  & 2                     \\ \hline
10                  & \multicolumn{1}{c|}{7}                  & \multicolumn{1}{c|}{5}                   & \multicolumn{1}{c|}{94}                         & 168                       & \multicolumn{1}{c|}{5}           & \multicolumn{1}{c|}{7}                & \multicolumn{1}{c|}{4}                & \multicolumn{1}{c|}{4}                          & \multicolumn{1}{c|}{6}                  & 7                     \\ \hline
\end{tabular}%

}
\caption{Empirical results of the preliminary user study $n=10$. We record two performance metrics, the number of blocks stacked and grams of rice scooped, and six UX metrics after the participants completed all tasks.}
\label{tab:results_all}
\end{table*}

%% file: includes/discussion.tex
\section{Discussion \& Conclusion}

In this paper, we proposed an AR-based puppeteering system that teleoperates a physical robot through manipulating a virtual robot in AR using a headset.
Our system follows the intuitive leader-follower puppeteering setup but cuts the hardware requirements by realizing the leader robot arm as a virtual robot. In addition, we can display the delay and status via an additional transparent rendering mirroring the real robot. One of the limitations of our framework is that the virtual robot does not display the same amount of inertia as a real physical robot does as the controller is significantly lighter to move than a real robot.

Besides the system development, we implemented an initial empirical pilot study, featuring ten participants where most possessed a profound robotics background and limited experience in AR. There were two tasks designed for the study and six UX metrics were measured. The results showed our system has the capability to finish practical robotics tasks like grasping and moving, by merely controlling the virtual robot and meanwhile puppeteering the physical robot. The UX metrics revealed that users held a positive attitude toward our system and disclosed potential interest in using it. Why did we obtain such results? First, we believe that our system is easy to learn, operate, and manipulate without any complex configuration steps. Secondly, we think our system ameliorated the safety issues considerably by only manipulating the virtual robot, compared to a physical puppeteering system involving direct interaction with a physical robot that could cause unwanted safety consequences. In addition, the immersion provided by our system to users is a crucial trait in achieving high user satisfaction and likeability.

In future work, we want to extend the system to a bimanual setup as well as extend it to arbitrary robot models.

%% file: IEEEexample.bib
@inproceedings{welle2024quest2ros,
  title={Quest2ROS: An App to Facilitate Teleoperating Robots},
  author={Welle, Michael C and Ingelhag, Nils and Lippi, Martina and Wozniak, Maciej and Gasparri, Andrea and Kragic, Danica},
  booktitle={7th International Workshop on Virtual, Augmented, and Mixed-Reality for Human-Robot Interactions},
  year={2024}
}


%% file: references.bib
@inproceedings{moletta2023virtual,
  title={A virtual reality framework for human-robot collaboration in cloth folding},
  author={Moletta, Marco and Wozniak, Maciej K and Welle, Michael C and Kragic, Danica},
  booktitle={2023 IEEE-RAS 22nd International Conference on Humanoid Robots (Humanoids)},
  pages={1--7},
  year={2023},
  organization={IEEE}
}

@article{ingelhag2024robotic,
  title={A Robotic Skill Learning System Built Upon Diffusion Policies and Foundation Models},
  author={Ingelhag, Nils and Munkeby, Jesper and van Haastregt, Jonne and Varava, Anastasia and Welle, Michael C and Kragic, Danica},
  journal={arXiv preprint arXiv:2403.16730},
  year={2024}
}

@article{chi2023diffusion,
  title={Diffusion policy: Visuomotor policy learning via action diffusion},
  author={Chi, Cheng and Feng, Siyuan and Du, Yilun and Xu, Zhenjia and Cousineau, Eric and Burchfiel, Benjamin and Song, Shuran},
  journal={arXiv preprint arXiv:2303.04137},
  year={2023}
}

@article{makhataeva2020augmented,
  title={Augmented reality for robotics: A review},
  author={Makhataeva, Zhanat and Varol, Huseyin Atakan},
  journal={Robotics},
  volume={9},
  number={2},
  pages={21},
  year={2020},
  publisher={MDPI}
}

@inproceedings{fu2024mobile,
  author    = {Fu, Zipeng and Zhao, Tony Z. and Finn, Chelsea},
  title     = {Mobile ALOHA: Learning Bimanual Mobile Manipulation with Low-Cost Whole-Body Teleoperation},
  booktitle = {arXiv},
  year      = {2024},
}

@inproceedings{zhang2023see,
  title={See or Hear? Exploring the Effect of Visual/Audio Hints and Gaze-assisted Instant Post-task Feedback for Visual Search Tasks in AR},
  author={Zhang, Yuchong and Nowak, Adam and Xuan, Yueming and Romanowski, Andrzej and Fjeld, Morten},
  booktitle={2023 IEEE International Symposium on Mixed and Augmented Reality (ISMAR)},
  pages={1113--1122},
  year={2023},
  organization={IEEE}
}

@inproceedings{zhang2023playing,
  title={Playing with data: An augmented reality approach to interact with visualizations of industrial process tomography},
  author={Zhang, Yuchong and Xuan, Yueming and Yadav, Rahul and Omrani, Adel and Fjeld, Morten},
  booktitle={IFIP Conference on Human-Computer Interaction},
  pages={123--144},
  year={2023},
  organization={Springer}
}

@article{zhang2021supporting,
  title={Supporting visualization analysis in industrial process tomography by using augmented reality—a case study of an industrial microwave drying system},
  author={Zhang, Yuchong and Omrani, Adel and Yadav, Rahul and Fjeld, Morten},
  journal={Sensors},
  volume={21},
  number={19},
  pages={6515},
  year={2021},
  publisher={MDPI}
}

@inproceedings{sakashita2017you,
  title={You as a puppet: evaluation of telepresence user interface for puppetry},
  author={Sakashita, Mose and Minagawa, Tatsuya and Koike, Amy and Suzuki, Ippei and Kawahara, Keisuke and Ochiai, Yoichi},
  booktitle={Proceedings of the 30th annual ACM symposium on user Interface software and technology},
  pages={217--228},
  year={2017}
}

@article{bejczy2020mixed,
  title={Mixed reality interface for improving mobile manipulator teleoperation in contamination critical applications},
  author={Bejczy, Bence and Bozyil, Rohat and Vai{\v{c}}ekauskas, Evaldas and Petersen, Sune Baag{\o} Krogh and B{\o}gh, Simon and Hjorth, Sebastian Schleisner and Hansen, Emil Blixt},
  journal={Procedia Manufacturing},
  volume={51},
  pages={620--626},
  year={2020},
  publisher={Elsevier}
}

@inproceedings{milgram1993applications,
  title={Applications of augmented reality for human-robot communication},
  author={Milgram, Paul and Zhai, Shumin and Drascic, David and Grodski, Julius},
  booktitle={Proceedings of 1993 IEEE/RSJ International Conference on Intelligent Robots and Systems (IROS'93)},
  volume={3},
  pages={1467--1472},
  year={1993},
  organization={IEEE}
}

@inproceedings{quintero2018robot,
  title={Robot programming through augmented trajectories in augmented reality},
  author={Quintero, Camilo Perez and Li, Sarah and Pan, Matthew KXJ and Chan, Wesley P and Van der Loos, HF Machiel and Croft, Elizabeth},
  booktitle={2018 IEEE/RSJ International Conference on Intelligent Robots and Systems (IROS)},
  pages={1838--1844},
  year={2018},
  organization={IEEE}
}

@inproceedings{mohareri2011autonomous,
  title={Autonomous humanoid robot navigation using augmented reality technique},
  author={Mohareri, Omid and Rad, Ahmad B},
  booktitle={2011 IEEE International Conference on Mechatronics},
  pages={463--468},
  year={2011},
  organization={IEEE}
}

@inproceedings{slyper2015mirror,
  title={Mirror puppeteering: Animating toy robots in front of a webcam},
  author={Slyper, Ronit and Hoffman, Guy and Shamir, Ariel},
  booktitle={Proceedings of the Ninth International Conference on Tangible, Embedded, and Embodied Interaction},
  pages={241--248},
  year={2015}
}

@mastersthesis{holland2018visual,
  title={Visual puppeteering using the Vizualeyez: 3D motion capture system},
  author={Holland, Mark},
  type={{B.S.} thesis},
  year={2018},
  school={University of Twente}
}

@inproceedings{aravind2015automated,
  title={Automated Puppetry—Robo-Puppet{\copyright}},
  author={Aravind, MA and Dinesh, NS and Rao, Nori Chalapathi and Charan, P Ram},
  booktitle={ICoRD’15--Research into Design Across Boundaries Volume 2: Creativity, Sustainability, DfX, Enabling Technologies, Management and Applications},
  pages={579--590},
  year={2015},
  organization={Springer}
}

@inproceedings{luvcny2023robot,
  title={Robot at the Mirror: Learning to Imitate via Associating Self-supervised Models},
  author={L{\'u}{\v{c}}ny, Andrej and Malinovsk{\'a}, Krist{\'\i}na and Farka{\v{s}}, Igor},
  booktitle={International Conference on Artificial Neural Networks},
  pages={471--482},
  year={2023},
  organization={Springer}
}

@article{rubagotti2022perceived,
  title={Perceived safety in physical human--robot interaction—A survey},
  author={Rubagotti, Matteo and Tusseyeva, Inara and Baltabayeva, Sara and Summers, Danna and Sandygulova, Anara},
  journal={Robotics and Autonomous Systems},
  volume={151},
  pages={104047},
  year={2022},
  publisher={Elsevier}
}

@article{hasan2007effects,
  title={Effects of interface style on user perceptions and behavioral intention to use computer systems},
  author={Hasan, Bassam and Ahmed, Mesbah U},
  journal={Computers in Human Behavior},
  volume={23},
  number={6},
  pages={3025--3037},
  year={2007},
  publisher={Elsevier}
}

@article{adams2009multiple,
  title={Multiple robot/single human interaction: Effects on perceived workload},
  author={Adams, Julie A},
  journal={Behaviour \& Information Technology},
  volume={28},
  number={2},
  pages={183--198},
  year={2009},
  publisher={Taylor \& Francis}
}

@article{bartneck2009measurement,
  title={Measurement instruments for the anthropomorphism, animacy, likeability, perceived intelligence, and perceived safety of robots},
  author={Bartneck, Christoph and Kuli{\'c}, Dana and Croft, Elizabeth and Zoghbi, Susana},
  journal={International journal of social robotics},
  volume={1},
  pages={71--81},
  year={2009},
  publisher={Springer}
}
